\documentclass[11pt,a4paper]{article}
\usepackage[hyperref]{acl2019}
\usepackage{times}
\usepackage{pgfplots}
\usepackage{multirow}
\usepackage{algorithm}
\usepackage{algpseudocode}
\usepackage{tabularx}
\usepackage{latexsym}
\usepackage[many]{tcolorbox}    	% for COLORED BOXES (tikz and xcolor included)
\newtcolorbox{boxK}{
    sharpish corners, % better drop shadow
    boxrule = 0pt,
    toprule = 4.5pt, % top rule weight
    enhanced,
    fuzzy shadow = {0pt}{-2pt}{-0.5pt}{0.5pt}{black!35} % {xshift}{yshift}{offset}{step}{options} 
}

\usepackage{url}

\aclfinalcopy % Uncomment this line for the final submission
\title{Improving the Natural Language Inference robustness to hard dataset by data augmentation and preprocessing}

\author{Zijiang YANG \\
  School of Computer Science  \\
  University of Texas \\ 
  \texttt{zy4957@eid.utexas.edu} }

\date{}

\algnewcommand\RETURN{\State \algorithmicreturn}%
\algnewcommand\PROCEDURE{\item[\algorithmicprocedure]}%
\algnewcommand{\algvar}[1]{{\text{\ttfamily\detokenize{#1}}}}
\algnewcommand{\algarg}[1]{{\text{\ttfamily\itshape\detokenize{#1}}}}

\begin{document}
\maketitle
\begin{abstract}
Natural Language Inference (NLI) is the task of inferring whether the hypothesis can be justified by the given premise.
Basically, we classify the hypothesis into three labels(entailment, neutrality and contradiction) given the premise. 
NLI was well studied by the previous researchers. A number of models, especially the transformer based ones, have achieved significant improvement on these tasks.
However, it is reported that these models are suffering when they are dealing with hard datasets. Particularly, they perform much worse when dealing with unseen out-of-distribution premise and hypothesis. They may not understand the semantic content but learn the spurious correlations. In this work, we propose the data augmentation and preprocessing methods to solve the word overlap, numerical reasoning and length mismatch problems. These methods are general methods that do not rely on the distribution of the testing data and they help improve the robustness of the models.
\end{abstract}

% \section{Credits}

% This document has been adapted from the instructions
% for earlier ACL and NAACL proceedings,
% including 
% those for 
% NAACL 2019 by Stephanie Lukin and Alla Roskovskaya, 
% ACL 2018 by Shay Cohen, Kevin Gimpel, and Wei Lu, 
% NAACL 2018 by Margaret Michell and Stephanie Lukin,
% 2017/2018 (NA)ACL bibtex suggestions from Jason Eisner,
% ACL 2017 by Dan Gildea and Min-Yen Kan, 
% NAACL 2017 by Margaret Mitchell, 
% ACL 2012 by Maggie Li and Michael White, 
% those from ACL 2010 by Jing-Shing Chang and Philipp Koehn, 
% those for ACL 2008 by JohannaD. Moore, Simone Teufel, James Allan, and Sadaoki Furui, 
% those for ACL 2005 by Hwee Tou Ng and Kemal Oflazer, 
% those for ACL 2002 by Eugene Charniak and Dekang Lin, 
% and earlier ACL and EACL formats.
% Those versions were written by several
% people, including John Chen, Henry S. Thompson and Donald
% Walker. Additional elements were taken from the formatting
% instructions of the \emph{International Joint Conference on Artificial
%   Intelligence} and the \emph{Conference on Computer Vision and
%   Pattern Recognition}.
 
\section{Introduction}

The Natural Language Inference (NLI) is the task of recognizing the textual entailment by determining the relationship between the hypothesis and the given premise. It is a classification task that returns the label \textit{entailment(0)/contradiction(2)} when the hypothesis agrees/disagrees with the premise. If the correctness of the hypothesis is unknown given the context of the premise, this task will return the label \textit{neutrality(1)}. The NLI task has been developed as a benchmark task for natural language understanding research\cite{giampiccolo-etal-2007-third},\cite{bowman-etal-2015-large}. The objective of this task is to let the natural language models understand the contextual semantics and make logical decisions. 

A lot of models have been proposed to solve this NLI task: BERT \cite{devlin-etal-2019-bert}, RoBERTa \cite{Liu2019RoBERTaAR}, XLNet \cite{Yang2019XLNetGA}. The most recent state-of-the-art model is the Masked Language Modeling(MLM) pre-taining models like BERT or its variants. These models understand the contextual relationships by masking out the token and reconstructing it back. It is shown to be effective in the downstream task like the NLI task. Recently the pre-trained model ELECTRA-small \cite{Clark2020ELECTRA:} gained much popularity due to its computational efficiency in solving the downstream task. 

These transformer-based models consistently achieve high accuracies on the simple datasets. However, such models may not really do the inference tasks. Rather, they may simply perform the 'pattern matching' between the testing data and the training data(\cite{Levesque2014OnOB}; \cite{Papernot2016PracticalBA}). One simple example would be the word overlap feature. The model may just compute the statistics of word overlap between the premise and hypothesis. The classification result will be entailment if the overlapping is significant. Such process is definitely not inference but memorization. To stress test and evaluate the capability of the models, more sophisticated datasets were generated by previous researchers. They deliberately generated data that tend to fool the model but not the human readers. The model performance on these hard datasets is not satisfactory with accuracies below $50\%$.

To address this problem, we analyze the limitations of current model on three aspects: (1) word overlap; (2) numerical reasoning; (3) length mismatch. Several methods are proposed to tackle these problems. 

The paper selects the ELECTRA-small as the base model for conducting the analysis. We will focus on its performance on the simple dataset SNLI and then evaluate the model on other hard datasets. The paper is structured as follows: Section 1 introduces the background information; Section 2 discusses the weak robustness of the model. Section 3 will analyze the associated problems and propose several improvement methods. The experimental results will be fully analyzed. Finally, we make the conclusion for the paper in Section 4.

% \begin{boxK}
% Misclassification of word overlap example \\
% Premise: The student supports the teacher. \\
% Hypothesis: The teacher supports the student. \\ 
% Gold label: Neutrality(1) \\
% Model prediction: Entailment(0)
% \end{boxK}

% \begin{boxK}
% Misclassification of word overlap example \\
% Premise: The student supports the teacher. \\
% Hypothesis: The teacher supports the student. \\ 
% Gold label: Neutrality(1) \\
% Model prediction: Entailment(0)
% \end{boxK}

% \begin{boxK}
% Misclassification of word overlap example \\
% Premise: The student supports the teacher. \\
% Hypothesis: The teacher supports the student. \\ 
% Gold label: Neutrality(1) \\
% Model prediction: Entailment(0)
% \end{boxK}

\begin{tcolorbox}[colback=red!5!white,colframe=red!75!black,title=Misclassification of word overlap example]
P: The student supports the teacher. \\
H: The teacher supports the student.
\tcblower
Gold label: Neutrality(1) \\
Model result: Entailment(0)
\end{tcolorbox}

\begin{tcolorbox}[colback=red!5!white,colframe=red!75!black,title=Misclassification of numerical reasoning example]
P: He (born in 1935) was an actor. \\
H: He was born before 1934.
\tcblower
Gold label: Contradiction(2) \\
Model result: Entailment(0)
\end{tcolorbox}

\begin{tcolorbox}[colback=red!5!white,colframe=red!75!black,title=Misclassification of length mismatched example]
P: Alan Metter is an American film director whose most notable credits include "Back to School" starring Rodney Dangerfield, and "Girls Just Want to Have Fun" with Sarah Jessica Parker. He also produced and directed the 1983 television special "The Winds of Whoopie" for Steve Martin. \\
H: "The Winds of Whoopie" was not produced to play in theaters.
\tcblower
Gold label: Entailment(0)  \\
Model result: Neutrality(1)
\end{tcolorbox}

\section{Weak Robustness of the NLI models}
Previous researcher have done a lot of work on the error analysis of the best-performing transformer model on different datasets (\cite{nie-bansal-2017-shortcut};\cite{naik-etal-2018-stress}). They have discovered the weak robustness of the models.  To analyze the robusness of the models, we trained the model(ELECTRA-small) on the benchmark dataset, SNLI, whose information can be found in the section 3.1. We discovered that the model cannot actually conduct the real inference even on the simplest form of premise-hypothesis pairs, if such pairs are in the out-of-distribution. Based on our analysis of the error types, we can conclude the followings are the main common mistake the model would make:
\begin{itemize}
    \item \textbf{Word overlap}: Large overlap between the premise and the hypothesis causes wrong classification decision. For the example on the text-box, we can that the premise \textit{'The student supports the teacher.'} and the hypothesis \textit{'The teacher supports the student.'} have 100\% word overlap. However, the model will predict the label entailment with probability close to $1$. Obviously, in this example, the model cannot learn the relationship between the subject and the object.

    \item \textbf{Numerical reasoning}: Another difficulty with the model is its numerical reasoning capability. When it comes to reasoning about the chronological orders of the events, it often gives very strange answers. Like the example in the text-box, it gives completely opposite label, contradiction for the gold label vs. entailment for the model result. Also, we find that the model cannot perform well on the numbers counting, comparison and basic arithmetic tasks. 
    % The reason may be the lack of the training data. 

    \item \textbf{Length mismatched}: The premise may be much longer than the hypothesis. The information may be hidden inside only one sentence in the premise, while the other sentences can be the distraction for the model. Take the example in the text-box. The whole paragraph is too long for the model to understand the location of \textit{'The Winds of Whoopie'}. The model gets puzzled and produces the label neutrality. However, if we remove the first sentence of the premise, the model result will be entailment. Moreover, if we extract the important information out from the premise and generate a new premise: \textit{'He produced the television special 'The Winds of Whoopie''}. The probability for predicting entailment will be much higher, which is close to 1.
    
\end{itemize}

\subsection{Discussion}
Based on the out-of-distribution examples, we can observe that the model exhibits weak robustness. The NLI models are prone to learning the superficial relationship between the premise and the hypothesis such as the statistics of the world overlapping ratio. The model tends to memorize but not infer the relationships. In the following sections, we would train the model to improve its robustness via data augmentation and data preprocessing.

\section{Improve the NLI Models' Robustness}

\subsection{Experimental Data}
We include various sets of data to test how the models can learn from the data and how well they can perform on the in-distribution and out-of-distribution data. The data used in this experiment include the following datasets. The difficulty is in the increasing order:
\begin{enumerate}
    \item SNLI \cite{bowman-etal-2015-large}: Stanford Natural Language Inference dataset is a widely used dataset for training and evaluating the NLI model. It contains a collection of 570,000 human-written English sentence pairs for balanced classification.
    
    \item MultiNLI \cite{MultiNLI}: Multi-Genre Natural Language Inference dataset is a crowd-sourced collection of 433,000 sentence pairs with annotation. It is an extension and more diversion version of the SNLI. While SNLI mainly focuses on the image captions, the MultiNLI captures a variety of genres including fiction, letters, speech, report and so on. 

    \item HANS \cite{DBLP:journals/corr/abs-1902-01007}: It is mainly for evaluating the NLI models on the specific heuristic but invalid hypotheses. These heuristics are likely to be learned by the model but there are not the real inference.

    \item ANLI \cite{nie2019adversarial}: It is designed to challenge the NLI models by introducing adversarial examples to the dataset. The data is collected via an iterative, adversarial human-model procedure. Its main goal is to try to fool the model with premise-hypothesis sentence pairs. It contains 3 rounds of datasets(r1, r2 and r3). r3 is built on top of r2 and r2 is built on top of r1. In terms of difficulty: $r3 > r2 > r1$
\end{enumerate}

\subsection{Methodology}
To improve the robustness of the NLI model on the above-mentioned three aspects. Different methods are proposed:

\subsubsection{Word Overlap}
We train the models via augmenting simple premise-hypothesis pairs that have large overlapping. We construct the additional data via the following steps: 
\begin{itemize}
    \item Extract out the words in the training corpus that have part-of-speech tag \textit{Noun(NN)} and name it by set $N$.
    \item Extract out the words in the training corpus that have part-of-speech tag \textit{Verb(V)} and name it by set $V$.
    \item Randomly pick two words $n1, n2$ from $N$ and one word $v1$ from $V$.
    \item Construct the additional dataset by generating 3 premises: 
        (1)\textit{The $n1$ $v1$ the $n2$.}
        (2)\textit{The $n1$ does not $v1$ the $n2$.}
        (3)\textit{The $n2$ $v1$ the $n1$.}
        And the hypothesis will be the \textit{The $n1$ $v1$ the $n2$}. So we basically construct 3 pairs with \textit{entailment/contradiction/neutrality} labels.
\end{itemize}

\subsubsection{Numerical Reasoning}
Improving numerical reasoning is a difficult task that includes many aspects. We will focus on improving the reasoning about the chronological order of the events. We solve it by augmenting simple premise-hypothesis pairs that ask for comparing the years. The pairs are generated by the following steps:
\begin{itemize}
    \item Randomly generate an integer $i1$ from 0 to 2020 for birth year.
    \item Randomly generate an integer $i2$ from $i1+1$ to $i1+100$ for death year.
    \item The premise will be: \textit{He ($i1$ - $i2$).}
    \item Randomly generate an integer $i3$ from  $i1-100$ to $i1+100$.
    \item The hypothesis will be:  \textit{He was born before $i3$.} The label will be entailment if $i3>i1$ else contradiction. 
    \item Similar hypothesis can be :  \textit{He was born after $i4$.}  \textit{He died before $i5$.}  \textit{He died after $i6$.} The labels are assgined accordingly.
\end{itemize}

\subsubsection{Length Mismatch}
Length mismatch problem is a built-in problem for the transformer type of models. The length of the input sentence is kept fixed before entering the model. Short sentences will be padded while the long sentences will be truncated. The situation is much worse for the long sentences as some useful and important information will be discarded during the training and evaluation process. Even if we adjust the designated input length for the model, it still cannot handle the long paragraphs. We advocate that for the paragraph input, it is advised to split the paragraph into sentences and then test the hypothesis for each sentence. If any sentence was found to be strongly agreeing or disagreeing the hypotheses, we return the entailment/contradiction label. Otherwise, we return the neutrality label. The details of the \textbf{Split} algorithm is shown in Alogorithm.\ref{alg:split}:

\begin{algorithm}
\caption{Split algorithm for paragraph premise}\label{alg:split}
\begin{algorithmic}
\Require Premise is a paragraph that contains multiple sentences.
% \Ensure $y = x^n$
\PROCEDURE{{classifyNLI}(\textit{Hypothesis}, \textit{Premise})}
\State $h \gets Hypothesis$
\State $pList \gets strSplit(Premise, '.')$
\State $c \gets 0.8$  \Comment{Set the cutoff probability to be 0.8}
\State $n \gets len(pList)$
\State $i \gets 0$
\While{$i < n$}
\State $V \gets Softmax(Model(pList[i], h))$
\If{$V[0] > c$} 
    \RETURN {} $0$   \Comment{Return Entailment}
\ElsIf{$V[2] > c$}
    \RETURN {} $2$   \Comment{Return Contradiction}
\EndIf
\State $i \gets i+1$ 
\EndWhile
\RETURN {} $1$   

\Comment{No strong Entailment or Contradiction sentence is found, return Neutrality}
\end{algorithmic}
\end{algorithm}

\subsection{Experimental Result and Analysis}
We first conduct the experiments on the original SNLI dataset. It is reported that the ELECTRA-small model trained on the SNLI training dataset achieves very high accuracy 0.882 in the SNLI testing dataset, as is shown in the table \ref{tab:my-table}. However, the model result is bad on the HANS dataset. The model seems to learn the heuristics but not the inference. Additionally, the model performs very poorly on hard dataset ANLI. The accuracy is as low as 0.3, which means the model is almost close to random guessing. 

By applying our proposed methods, we first observe a significant accuracy increase on the HANS dataset as we construct the counter-heuristic examples to train the model. The increase is around 14\% for data augmentation and 12\% for the data augmentation with Split. It is worth noting that the performance for the SNLI dataset drops by just 3\% by learning some out-of-distribution data. 

For the model results on the hard datasets ANLI, we see the improvement is substantial. With the data augmentation, the average improvement on ANLI r1,r2,r3 dataset is 3\%. With the Split algorithm, the average improvement reaches 5\%. By applying both techniques, the average improvment reaches 7\%. We see the potential of the model on the hard dataset with the proposed methods. Even though the model is only trained on the simple dataset, the model performance is comparable to the one trained on the hard datasets. The generalization capability and robustness of the model is largely improved.

Furthermore, we test our methods by varying the number of augmented examples to the original datasets. The results are shown in the figures below. The first observation will be the obvious accuracy increase on the hard dataset ANLI regardless of the number of augmented samples. The second observation will be the substantial accuracy improvement on HANS by augmenting more samples. Lastly, we observe that the performance improvement reaches the limit at the 1000 augmented samples. Therefore, with only 1000 additional sample, as is compared to 40,000 original training data size, it is sufficient for the proposed method to perform well.

\begin{figure}
        \centering
        \resizebox{\columnwidth}{!}{%
        \begin{tikzpicture}
        \begin{axis}[
            title={ Performance of ELECTRA-small 
            },
            xlabel={No. of samples augmented(x1000)},
            ylabel={Accuracy},
            xmin=0, xmax=5,
            ymin=0.25, ymax=0.45,
            xtick={0,0.5,1,2,3,4,5},
            ytick={0.25,0.30,0.35,0.40,0.45},
            legend pos=north west,
            ymajorgrids=true,
            grid style=dashed,
        ]
        \addplot[
            color=blue,
            mark=square,
            ]
            coordinates { 
            (0,0.306)(0.1,0.302)(0.25,0.314)(0.5,0.313)(0.75,0.313)
            (1,0.307)(2,0.326)(3,0.332)(4,0.329)(5,0.329)
            };    
        \addplot[
            color=red,
            mark=square,
            ]
            coordinates {  
            (0,0.345)(0.1,0.355)(0.25,0.354)(0.5,0.352)(0.75,0.366)
            (1,0.361)(2,0.368)(3,0.364)(4,0.370)(5,0.367)
            };
            
        \legend{ANLI\_r1, ANLI\_r1+Split}
        \end{axis}
        \end{tikzpicture}
        }
\end{figure}

\begin{figure}
        \centering
        \resizebox{\columnwidth}{!}{%
        \begin{tikzpicture}
        \begin{axis}[
            title={ Performance of ELECTRA-small 
            },
            xlabel={No. of samples augmented(x1000)},
            ylabel={Accuracy},
            xmin=0, xmax=5,
            ymin=0.25, ymax=0.45,
            xtick={0,0.5,1,2,3,4,5},
            ytick={0.25,0.30,0.35,0.40,0.45},
            legend pos=north west,
            ymajorgrids=true,
            grid style=dashed,
        ]
        \addplot[
            color=blue,
            mark=square,
            ]
            coordinates { 
            (0,0.308)(0.1,0.309)(0.25,0.311)(0.5,0.312)(0.75,0.318)
            (1,0.311)(2,0.323)(3,0.339)(4,0.325)(5,0.324)
            };    
        \addplot[
            color=red,
            mark=square,
            ]
            coordinates {  
            (0,0.359)(0.1,0.365)(0.25,0.374)(0.5,0.373)(0.75,0.389)
            (1,0.38)(2,0.379)(3,0.392)(4,0.389)(5,0.392)
            };
            
        \legend{ANLI\_r2, ANLI\_r2+Split}
        \end{axis}
        \end{tikzpicture}
        }
\end{figure}

\begin{figure}
        \centering
        \resizebox{\columnwidth}{!}{%
        \begin{tikzpicture}
        \begin{axis}[
            title={ Performance of ELECTRA-small 
            },
            xlabel={No. of samples augmented(x1000)},
            ylabel={Accuracy},
            xmin=0, xmax=5,
            ymin=0.25, ymax=0.45,
            xtick={0,0.5,1,2,3,4,5},
            ytick={0.25,0.30,0.35,0.40,0.45},
            legend pos=north west,
            ymajorgrids=true,
            grid style=dashed,
        ]
        \addplot[
            color=blue,
            mark=square,
            ]
            coordinates { 
            (0,0.30)(0.1,0.2975)(0.25,0.2942)(0.5,0.301)(0.75,0.3075)
            (1,0.30167)(2,0.3033)(3,0.311)(4,0.3141)(5,0.309)
            };    
        \addplot[
            color=red,
            mark=square,
            ]
            coordinates {  
            (0,0.369)(0.1,0.36167)(0.25,0.365)(0.5,0.361)(0.75,0.361)
            (1,0.3625)(2,0.3766)(3,0.37166)(4,0.3742)(5,0.3675)
            };
            
        \legend{ANLI\_r3, ANLI\_r3+Split}
        \end{axis}
        \end{tikzpicture}
        }
\end{figure}

\begin{figure}
        \centering
        \resizebox{\columnwidth}{!}{%
        \begin{tikzpicture}
        \begin{axis}[
            title={ Performance of ELECTRA-small 
            },
            xlabel={No. of samples augmented(x1000)},
            ylabel={Accuracy},
            xmin=0, xmax=5,
            ymin=0.45, ymax=0.70,
            xtick={0,0.5,1,2,3,4,5},
            ytick={0.40,0.450,0.5,0.550,0.60,0.65,0.7},
            legend pos=north west,
            ymajorgrids=true,
            grid style=dashed,
        ]
        \addplot[
            color=blue,
            mark=square,
            ]
            coordinates { 
            (0,0.491)(0.1,0.5134)(0.25,0.6171)(0.5,0.58523)(0.75,0.5609)
            (1,0.6243)(2,0.6263)(3,0.63746)(4,0.62253)(5,0.631)
            };    
        \addplot[
            color=red,
            mark=square,
            ]
            coordinates {  
            (0,0.494)(0.1,0.4856)(0.25,0.5567)(0.5,0.6048)(0.75,0.61446)
            (1,0.6037)(2,0.603)(3,0.616)(4,0.61463)(5,0.6132)
            };
            
        \legend{HANS, HANS+Split}
        \end{axis}
        \end{tikzpicture}
        }
\end{figure}

\begin{figure}
        \centering
        \resizebox{\columnwidth}{!}{%
        \begin{tikzpicture}
        \begin{axis}[
            title={ Performance of ELECTRA-small 
            },
            xlabel={No. of samples augmented(x1000)},
            ylabel={Accuracy},
            xmin=0, xmax=5,
            ymin=0.45, ymax=0.70,
            xtick={0,0.5,1,2,3,4,5},
            ytick={0.40,0.450,0.5,0.550,0.60,0.65,0.7},
            legend pos=south west,
            ymajorgrids=true,
            grid style=dashed,
        ]
        \addplot[
            color=blue,
            mark=square,
            ]
            coordinates { 
            (0,0.706)(0.1,0.68925)(0.25,0.6680)(0.5,0.63790)(0.75,0.61874)
            (1,0.64228)(2,0.663066)(3,0.66021)(4,0.663066)(5,0.66194)
            };    
        \addplot[
            color=red,
            mark=square,
            ]
            coordinates {  
            (0,0.623)(0.1,0.6095)(0.25,0.6146)(0.5,0.61395)(0.75,0.6114)
            (1,0.61558)(2,0.594294)(3,0.58115)(4,0.59195)(5,0.5892)
            };
            
        \legend{MultiNLI, MultiNLI+Split}
        \end{axis}
        \end{tikzpicture}
        }
\end{figure}

\begin{figure}
        \centering
        \resizebox{\columnwidth}{!}{%
        \begin{tikzpicture}
        \begin{axis}[
            title={ Performance of ELECTRA-small 
            },
            xlabel={No. of samples augmented(x1000)},
            ylabel={Accuracy},
            xmin=0, xmax=5,
            ymin=0.7, ymax=0.95,
            xtick={0,0.5,1,2,3,4,5},
            ytick={0.70, 0.75,0.80,0.85,0.900},
            legend pos=south west,
            ymajorgrids=true,
            grid style=dashed,
        ]
        \addplot[
            color=blue,
            mark=square,
            ]
            coordinates { 
            (0,0.882)(0.1,0.8753)(0.25,0.8643)(0.5,0.8499)(0.75,0.8379)
            (1,0.8499)(2,0.8591)(3,0.853)(4,0.8571)(5,0.8557)
            };    
        \addplot[
            color=red,
            mark=square,
            ]
            coordinates {  
            (0,0.833)(0.1,0.8414)(0.25,0.8453)(0.5,0.85)(0.75,0.8495)
            (1,0.85)(2,0.85)(3,0.8504)(4,0.8527)(5,0.8527)
            };
            
        \legend{SNLI, SNLI+Split}
        \end{axis}
        \end{tikzpicture}
        }
\end{figure}

% Please add the following required packages to your document preamble:

\begin{table*}[th!]
\centering

    \begin{tabular}{lllllllll}
     \hline
    \multirow{2}{*}{Trained on} & \multirow{2}{*}{\begin{tabular}[c]{@{}l@{}}Preprocess\\ technique\end{tabular}} & \multicolumn{3}{c}{ANLI}                                                 & \multicolumn{1}{c}{\multirow{2}{*}{HANS}} & \multicolumn{1}{c}{\multirow{2}{*}{MultiNLI }} & \multicolumn{1}{c}{\multirow{2}{*}{SNLI}} & \multirow{2}{*}{Average} \\
                                &                                                                                 & \multicolumn{1}{c}{r1} & \multicolumn{1}{c}{r2} & \multicolumn{1}{c}{r3} & \multicolumn{1}{c}{}                      & \multicolumn{1}{c}{}                      & \multicolumn{1}{c}{}                      &                          \\ \hline 

    \textit{Baseline(simple dataset)} \\ 
    SNLI                        &                                                                                 & 0.306                  & 0.308                  & 0.300                  & 0.491                                     & 0.706                                     & 0.882                                     & 0.499                    \\
    SNLI                        & +Split                                                                          & 0.345                  & 0.359                  & 0.369                  & 0.494                                     & 0.623                                     & 0.833                                     & 0.502                    \\ 
    \hline

    \textit{Proposed} \\ 

        SNLI(DataAug)               &                                                                                 & 0.332                  & 0.339                  & 0.311                  & 0.637                                     & 0.660                                     & 0.853                                     & 0.522                    \\
    SNLI(DataAug)               & +Split                                                                          & 0.359                  & 0.392                  & 0.371                  & 0.616                                     & 0.581                                     & 0.851                                     & 0.528                    \\
\hline

    \textit{Baseline(hard dataset)} \\ 
    
    ANLI\_r1                    &                                                                                 & 0.373                  & 0.357                  & 0.358                  & 0.492                                     & 0.425                                     & 0.465                                     & 0.412                    \\
    ANLI\_r1                    & +Split                                                                          & 0.396                  & 0.377                  & 0.359                  & 0.486                                     & 0.437                                     & 0.443                                     & 0.416                    \\
    ANLI\_r2                    &                                                                                 & 0.460                  & 0.404                  & 0.360                  & 0.489                                     & 0.414                                     & 0.377                                     & 0.417                    \\
    ANLI\_r2                    & +Split                                                                          & 0.434                  & 0.403                  & 0.358                  & 0.479                                     & 0.383                                     & 0.362                                     & 0.403                    \\
    ANLI\_r3                    &                                                                                 & 0.471                  & 0.392                  & 0.406                  & 0.398                                     & 0.615                                     & 0.581                                     & 0.477                    \\
    ANLI\_r3                    & +Split                                                                          & 0.473                  & 0.422                  & 0.395                  & 0.376                                     & 0.600                                     & 0.591                                     & 0.476                    \\
    \hline
    
    \textit{Baseline(simple + hard)} \\ 
    SNLI+ANLI\_r1               &                                                                                 & 0.413                  & 0.336                  & 0.342                  & 0.488                                     & 0.686                                     & 0.861                                     & 0.521                    \\
    SNLI+ANLI\_r1               & +Split                                                                          & 0.374                  & 0.380                  & 0.356                  & 0.498                                     & 0.636                                     & 0.846                                     & 0.515                    \\
    SNLI+ANLI\_r2               &                                                                                 & 0.480                  & 0.387                  & 0.360                  & 0.490                                     & 0.680                                     & 0.831                                     & 0.538                    \\
    SNLI+ANLI\_r2               & +Split                                                                          & 0.416                  & 0.381                  & 0.376                  & 0.493                                     & 0.674                                     & 0.843                                     & 0.531                    \\
    SNLI+ANLI\_r3               &                                                                                 & 0.460                  & 0.396                  & 0.421                  & 0.490                                     & 0.724                                     & 0.830                                     & 0.553                    \\
    SNLI+ANLI\_r3               & +Split                                                                          & 0.432                  & 0.395                  & 0.403                  & 0.499                                     & 0.689                                     & 0.814                                     & 0.539                  \\ 
    \hline
    \end{tabular}

\caption{Label classification accuracy for models trained by different training data. They are tested on various datasets with different difficulty.}
\label{tab:my-table}
\end{table*}

% \subsection{Analysis}
% \subsection{Discussion}

\section{Conclusion}
In this work, we present a simple but effective data augmentation and preprocessing method to improve the generalization and robustness of the NLI model. Such method improves the model by solving the following three problems: (1) word overlap; (2) numerical reasoning; (3) length mismatched. The experimental results show that the model achieves more than 12\% on the HANS dataset and  6\% to 9\% on the ANLI dataset. The method helps prevent the heuristic search between the premise and hypothesis. Also, the preprocessing technique(Split) sheds the light on classifying the hypothesis with the paragraph-level premise while we only train the model on the sentence-level premises. The method can be effective by adding only 3\% of the original training dataset.

\bibliography{acl2019}
\bibliographystyle{acl_natbib}

\end{document}